\documentclass{Interspeech2024}

\interspeechcameraready

\title{Identifying Speakers in Dialogue Transcripts:\\A Text-based Approach Using Pretrained Language Models}

\name[affiliation={1,2,*}]{Minh}{Nguyen}
\name[affiliation={3}]{Franck}{Dernoncourt}
\name[affiliation={3}]{Seunghyun}{Yoon}
\name[affiliation={3}]{Hanieh}{Deilamsalehy}
\name[affiliation={3}]{Hao}{Tan}
\name[affiliation={3}]{Ryan}{Rossi}
\name[affiliation={3}]{Quan Hung}{Tran}
\name[affiliation={3}]{Trung}{Bui}
\name[affiliation={1}]{Thien Huu}{Nguyen}

\address{
  $^1$Department of Computer Science, University of Oregon
  $^2$AWS AI Labs
  $^3$Adobe Research\\
  $^*$\textit{\small Work done while the author was an intern at Adobe Research before joining AWS AI Labs}}
\email{minhnv@\{cs.uoregon.edu,amazon.com\},\{franck.dernoncourt,syoon,deilamsa,hatan,ryrossi,quanthdhcn,bui\}@adobe.com, thien@cs.uoregon.edu}


\usepackage{times}
\usepackage{latexsym}
\usepackage{amsmath,amssymb}
\usepackage{mathtools}
\usepackage{multirow}
\usepackage{graphicx}

\usepackage[utf8]{inputenc}

\usepackage{microtype}

\usepackage{inconsolata}

\begin{document}
\maketitle

\begin{abstract}
    We introduce an approach to identifying speaker names in dialogue transcripts, a crucial task for enhancing content accessibility and searchability in digital media archives. Despite the advancements in speech recognition, the task of text-based speaker identification (SpeakerID) has received limited attention, lacking large-scale, diverse datasets for effective model training. Addressing these gaps, we present a novel, large-scale dataset derived from the MediaSum corpus, encompassing transcripts from a wide range of media sources. We propose novel transformer-based models tailored for SpeakerID, leveraging contextual cues within dialogues to accurately attribute speaker names. Through extensive experiments, our best model achieves a great precision of 80.3\%, setting a new benchmark for SpeakerID. The data and code are publicly available here: \url{https://github.com/adobe-research/speaker-identification}
    
\end{abstract}

\section{Introduction}

The rapid expansion of dialogue-centric content across various media platforms, including television programs, online meetings, and radio podcasts, has significantly heightened user interest in accessing and exploring these rich sources of information and entertainment. In response to this growing demand, leading archival platforms and organizations such as YouTube, France's National Audiovisual Institute, and the British Broadcasting Corporation have dedicated considerable efforts towards the efficient storage and indexing of such content, facilitating its retrieval \cite{salmon2014effortless,vallet2016speech,burgess2018youtube}. Within this context, the challenge of accurately identifying speakers within dialogues—a process known as Speaker Identification (SpeakerID)—has emerged as a pivotal area of research. SpeakerID involves the task of recognizing and distinguishing between the voices of different speakers within an audio or video segment, aiming to assign the correct speaker names to each spoken segment. This process is crucial for enhancing the accessibility and searchability of multimedia content, enabling users to find segments featuring specific speakers. As a result, the development of effective SpeakerID systems has attracted significant research efforts, as evidenced by a body of work \cite{tranter2006really,poignant2014unsupervised,poignant2015multimodal,vallet-etal-2016-speech,le2017towards}, striving to overcome the challenges associated with this complex task.

\begin{table}[h]
\small
\centering
\resizebox{0.45\textwidth}{!}{
\begin{tabular}{|l|l|}
\hline
\textbf{Dialogue Transcript}                                & \textbf{Speaker} \\ \hline
"Good morning, everyone. This is John speaking."      & John                        \\ \hline
"Hi John, this is Sarah. Thanks for organizing this." & Sarah                       \\ \hline
"Absolutely, Sarah. And I think Mike has a question." & John                        \\ \hline
"Yes, I do. What's the timeline for our project?"     & Mike                        \\ \hline
\end{tabular}
}
\caption{An example for text-based Speaker ID.}
\label{tab:SpeakerIDExample}
\end{table}

Historically, SpeakerID research has predominantly focused on multimodal approaches, relying on both video/images and transcripts. While powerful, these multimodal systems demand substantial infrastructure support and struggle to handle audio-only dialogues. Recognizing these limitations, some researchers have shifted towards text-based SpeakerID, leveraging dialogue transcripts to identify speaker names, where a model needs to identify names for speakers in a given dialogue transcript \cite{tranter2006really,esteve2007extracting,jousse2009automatic,kuchavrova2014study}. An example for the task is presented in Table \ref{tab:SpeakerIDExample}. In this example, the task involves analyzing the dialogue to identify where speakers introduce themselves or are mentioned by name, and then attributing those segments of dialogue to the correct individuals. The shift to the text-based SpeakerID has been bolstered by advances in speech recognition technologies \cite{deng2014ensemble,zhang2018deep,kamath2019deep,wang-etal-2020-fairseq} and the emergence of pre-trained language models (PLMs) \cite{devlin-etal-2019-bert,liu2019roberta}, making the task increasingly viable. However, a notable gap remains: to our knowledge, no existing text-based SpeakerID research has utilized deep learning techniques or PLMs, and the scarcity of large-scale training datasets has further hindered model development. 

\begin{figure*}
    \centering
    \includegraphics[scale=0.48]{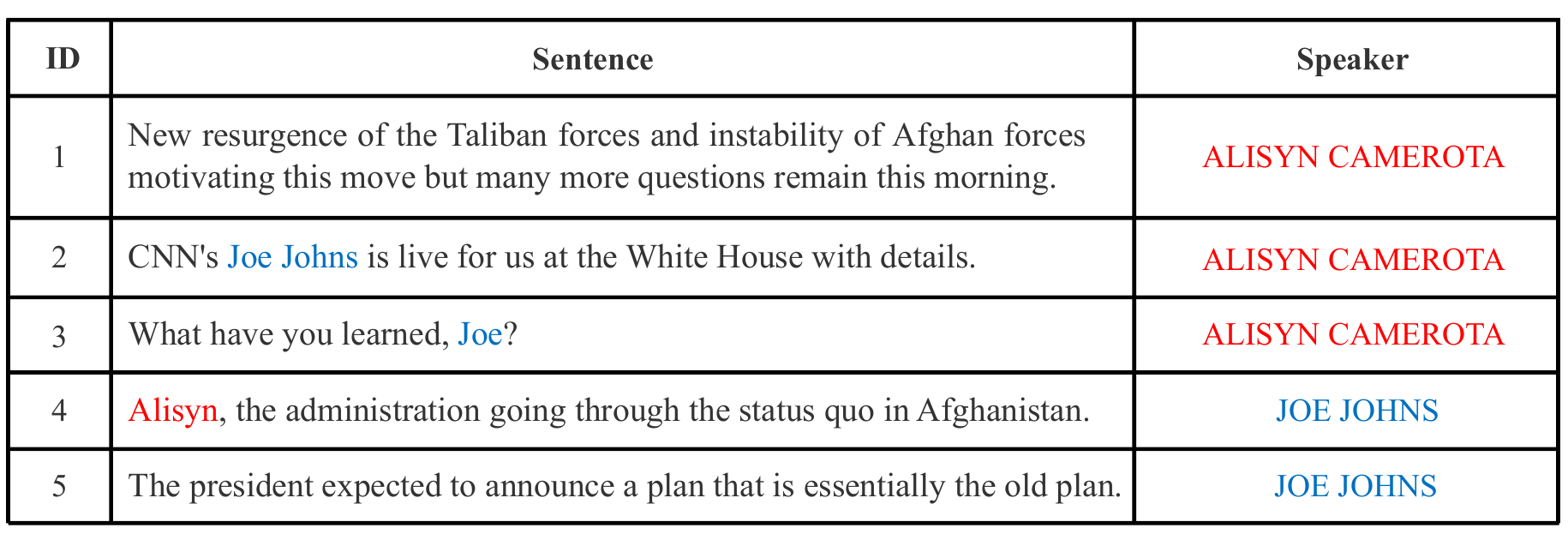}
    \caption{An example in the MediaSum dataset. In the SpeakerID setting, the speakers are not provided with their names at test time but their speaker identities such as ``speaker1'', ``speaker2'' produced by a speaker diarization system. A model performing SpeakerID needs to recover the actual names for the speakers based on the transcript.}
    \label{fig:example}
\end{figure*}

To address such issues, we first propose a simple method to automatically obtain a high-quality large-scale training data for text-based SpeakerID from the popular MediaSum corpus \cite{zhu-etal-2021-mediasum}, which contains transcripts for $463.6$K media interviews from the National Public Radio (NPR) and the Cable News Network (CNN). In addition to the transcripts, each interview comes with the information of the involved speakers as shown in Figure \ref{fig:example}. However, it is impossible for a text-based SpeakerID model to produce full names for the speakers (e.g., ``ALISYN CAMEROTA''); the model needs to assign variants of the names (e.g., ``Alisyn'') mentioned in the transcript to the speakers. As such, we propose to perform text matching to find mentions of speaker names in a transcript and use such names as labels for the task.

Furthermore, we propose novel transformer-based models for text-based SpeakerID. Our key observation is that speakers are often around when their names are mentioned during a meeting, i.e., it is often the case that we can assign a person name to the previous, current, or the next speaker. As such, we propose to represent the three possible speakers using their spoken sentences closest to the current sentence/utterance mentioning the person's name. In this way, we aim to find the correct speaker for a given name by pairing the name with each speaker. This is done by encoding the spoken sentences using a PLM and check if the given name belongs to any of them. To handle the case that multiple names could co-occur in the current sentence, we also explore another design for our model using Graph Convolutional Networks \cite{kipf2017semisupervised} to capture the relationships between the names (e.g., different names should be assigned to different speakers). 

To conduct experiments, we randomly sample a portion of our synthetic data generated from MediaSum and randomly split the data into train/dev/test sets. Experimental results show that our best model achieves a great performance with a precision of $80.3\%$ on the test set, demonstrating the quality of our proposed dataset and models for text-based SpeakerID. For the sake of simplicity, we will refer to text-based Speaker Identification as simply "SpeakerID" from this point forward.

\section{Methodology}
\subsection{Problem Definition}
We formalize the text-based speaker identification (SpeakerID) as follows. Given a dialogue transcript with anonymized speaker identities (e.g., ``speaker1'', ``speaker2'') and person names (e.g., ``Paul Erickson'') for each sentence in the transcript, find the actual names for each anonymized speaker identity. Here, the transcript can be obtained for the dialogue by a speech-to-text system such as Fairseq S2T \cite{wang-etal-2020-fairseq}, the speaker identities for each sentence can be produced by a speaker diarization system such as SOND \cite{du-etal-2022-speaker}, the person names can be detected by a named entity recognition (NER) such as Trankit \cite{nguyen-etal-2021-trankit}. In this work, we assume such information is available for our SpeakerID models.

\subsection{Data Collection}
\noindent \textbf{MediaSum Dataset}: MediaSum is a large-scale dialogue summarization dataset that was created by \cite{zhu-etal-2021-mediasum}. It contains $463.6$K interview-summary pairs from diverse news sources such as National Public Radio (NPR)\footnote{www.npr.org} and Cable News Network (CNN)\footnote{www.cnn.com}. The interviews cover a wide range of topics/domains, including politics, entertainment, sports, and technologies. In addition, each utterance/sentence in the transcript is tagged with the information of the speaker, including their names, titles, and affiliations. An example of the Mediasum transcripts is shown in Figure \ref{fig:example}.

\noindent \textbf{Data processing}: Given a MediaSum example, we perform the following steps to obtain training examples for SpeakerID:

\begin{itemize}
    \item \textbf{Step 1}: Detect person names in the transcript.
    \item \textbf{Step 2}: Anonymize the speaker information by replacing their actual names with speaker identities such as ``speaker1'', ``speaker2''.
    \item \textbf{Step 3}: Map the detected person names to the speaker identities via text matching. The names that do not match any speaker are assigned to a special speaker identity ``null''.
\end{itemize}
In this process, we use the state-of-the-art NER model from Trankit \cite{nguyen-etal-2021-trankit}, which achieves the state-of-the-art NER performance of 92.5 F1 on CoNLL English test set, to find spans for named entities in each sentence. Entities with the tag ``PERSON'' are considered as person names. The start and end tokens for each person name are then stored for the example. For the text matching between the person names and speaker actual names, we employ the Levenshtein Distance  \cite{yujian2007normalized} to perform the fuzzy text matching. In particular, Levenshtein Distance is a method for measuring the similarity between two strings of characters. The method computes the minimum changes (i.e., insertions, deletions, or substitutions of individual characters) needed to transform one string into the other. The similarity between the two strings can be measured as: $\theta = \frac{l_{sum} - d}{l_{sum}}$, where $l_{sum}$ is the total length of the two strings and $d$ is the computed Levenshtein distance. As names of the speakers can vary slightly in the transcript (e.g., missing last name), we find out that names with the similarity of at least $0.8$ can be effectively considered the same.

\subsection{Proposed Models}

\begin{figure}[ht]
    \centering
    \includegraphics[scale=0.57]{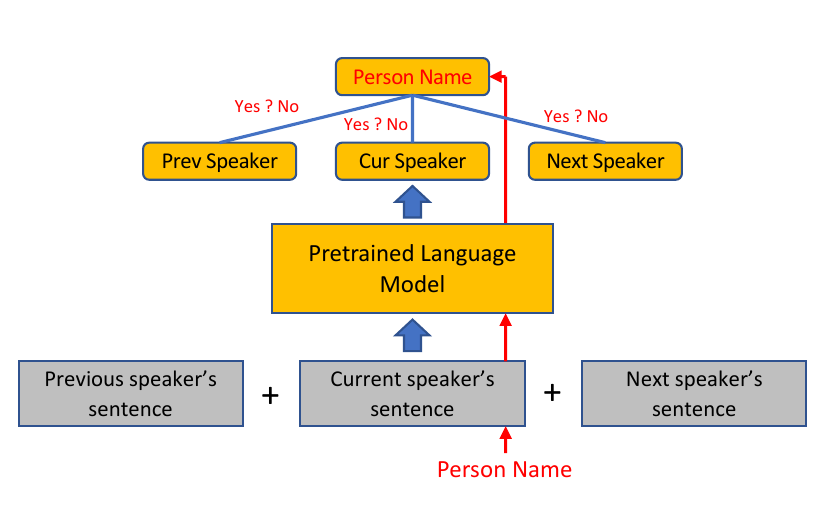}
    \caption{Overview of our proposed single-name model for SpeakerID.}
    \label{fig:single-name}
\end{figure}

\subsubsection{Single-Name Model}
\label{sec:single-name}
We present our first model design for SpeakerID in Figure \ref{fig:single-name}. In this model, we focus on a given person name $n$. For each occurrence of $n$, we identify the sentence or utterance $w_{cur}$ in which $n$ appears. This sentence is produced by a speaker, denoted as $s_{cur}$ (i.e., current speaker). We also consider the immediate dialogue context by identifying the sentences preceding and following $w_{cur}$, labeled as $w_{prev}$ and $w_{next}$, along with their corresponding speakers $s_{prev}$ and $s_{next}$ (previous and next speakers), respectively. This contextual framing is essential for understanding the dynamics of dialogue and speaker relationships.

To construct a comprehensive input sequence, we concatenate $w_{prev}$, $w_{cur}$, and $w_{next}$ into a single sequence $w$. In cases where either $w_{prev}$ or $w_{next}$ is missing (e.g., if $w_{cur}$ is at the beginning or end of a dialogue), we introduce padding sentences to maintain consistency in sequence formation. This concatenated sequence represents a broader dialogue context that encapsulates not just the mentioned name but also the surrounding conversational flow.

Upon forming this input sequence, we process it through a pretrained language model (PLM), such as RoBERTa \cite{liu2019roberta}, renowned for its ability to derive deep contextualized representations of text. By passing $w$ through the PLM, we extract the last-layer subword representations, which capture nuanced semantic and syntactic features of the text. To obtain word-level representations from these subwords, we average the subword representations for each word. Subsequently, to represent each sentence within our concatenated sequence, we compute the average of its word representations, yielding distinct vectors that encapsulate the essence of each sentence.

Given that these sentences originate from three different speakers, we posit that their vectors contain unique semantic signatures reflective of each speaker's communicative style or content. Thus, we treat these sentence vectors as proxies for the speakers themselves, assigning them as $\textbf{r}_{prev}$, $\textbf{r}_{cur}$, and $\textbf{r}_{next}$ for the previous, current, and next speakers, respectively.

For the person name $n$, its representation $\textbf{r}_{n}$ is derived by averaging the word representations within the span of $n$. This name vector, encapsulating the linguistic context of the name within the dialogue, is then paired with each speaker vector ($\textbf{r}_{prev}$, $\textbf{r}_{cur}$, $\textbf{r}_{next}$). These pairs form the basis for predicting the association between $n$ and the potential speakers.

Each of these concatenated pair vectors is inputted into a feed-forward neural network culminating in a sigmoid output layer, which outputs probability scores $p_{prev}$, $p_{cur}$, $p_{next}$ representing the likelihood of the name $n$ being associated with the previous, current, and next speakers, respectively. The model's learning objective is to minimize the standard cross-entropy loss between these predicted probabilities and the true speaker identities. This training process fine-tunes the model to discern the subtle cues within the dialogue that indicate speaker identities, thereby enhancing its ability to accurately attribute names to the correct speakers within complex conversational contexts.

\begin{figure}[h]
    \centering
    \includegraphics[scale=0.57]{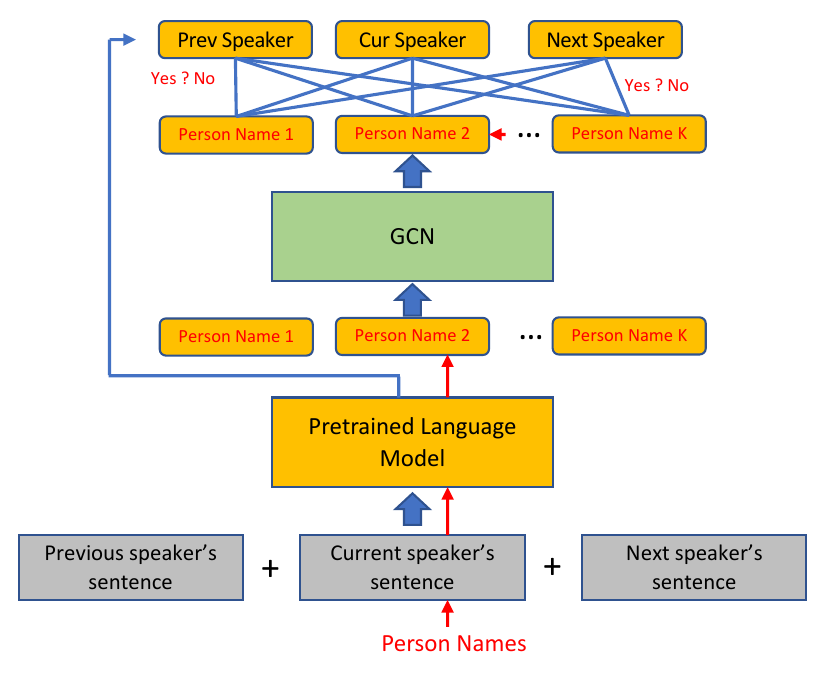}
    \caption{Overview of our proposed multi-name model for SpeakerID.}
    \label{fig:multi-name}
\end{figure}

\subsubsection{Multi-Name Model}

In scenarios where a sentence includes multiple names, our observations indicate that these names generally correspond to distinct speaker identities. This includes the possibility of a "null" speaker identity, which is used to represent instances where the speaker's identity may not be directly linked to any mentioned names within the dialogue. To address this complexity, we introduce a sophisticated model designed to simultaneously predict speaker identities for multiple names within a single sentence. This approach is particularly useful in dialogues where multiple individuals are referenced, necessitating a nuanced understanding of speaker identity.

Consider a sentence that mentions $K$ person names, represented by the vectors $\textbf{r}_1$, $\textbf{r}_2,\ldots, \textbf{r}_K$. In our model, each of these name representations is treated as a node within a fully connected graph $G$. The edges of this graph serve to represent the similarity between pairs of names, suggesting that names with higher similarity scores might share or be closely related to specific speaker identities.

To quantify the similarity between any two names, we employ the cosine similarity measure. Specifically, the weight of the edge between any two nodes (names) $i$ and $j$ in the graph is:
\begin{equation}
\alpha_{ij} = \textrm{softmax}(\frac{\textbf{r}^T_i \textbf{r}j}{\sum{j'} \textbf{r}^T_i \textbf{r}_{j'}})
\end{equation}
This formula essentially normalizes the cosine similarities between a name $i$ and all other names $j$, ensuring that the edge weights are comparable across the graph and facilitating a probabilistic interpretation of name similarity.

Upon establishing the graph structure with calculated edge weights, we proceed to employ a Graph Convolutional Network (GCN) \cite{kipf2017semisupervised} to refine the representations of each name. The GCN operates over $L$ layers, where each layer enhances the name representations by aggregating information from connected nodes (i.e., other names) weighted by their similarities. The operation at each layer $l$ is defined as:
\begin{equation}
\textbf{h}^l_i = \textrm{ReLU}(\sum^K_{j=1} \alpha_{ij}\textbf{W}^l\textbf{h}^{l-1}_j + \textbf{b}^l)
\end{equation}
In this equation, $\textbf{W}^l$ and $\textbf{b}^l$ represent the learnable weight matrix and bias for the $l$th layer of the GCN, respectively, and $\textbf{h}^0_i$ is the initial input representation for the name $i$.

This method allows the model to iteratively refine the representation of each name by incorporating contextual information from other names within the same sentence, effectively capturing the relational dynamics between mentioned individuals. Ultimately, the enhanced name representations can be paired with speaker representations, as detailed in a previous section, to accurately predict the corresponding speaker identities. This innovative approach leverages the power of GCN to understand and model the complex interrelations between multiple names mentioned in dialogues, offering a promising avenue for advancing the accuracy of speaker identification in rich, multimodal content.

\subsubsection{Inference}
At test time, there might be multiple names assigned to the same speaker identity. To make final predictions for the names, we simply select the name with the highest probability score.

\section{Experiments}
\subsection{Dataset} We randomly sample 200 meetings from the MediaSum dataset \cite{zhu-etal-2021-mediasum} in English language to create a SpeakerID dataset for experimental purpose. We randomly split the resulting dataset into train/dev/test with a ratio of 8/1/1. Statistics for the experimental dataset is shown in Table \ref{tab:datasets}.

\begin{table}[ht]
\small 
\begin{tabular}{|l|l|l|l|l|}
\hline
\textbf{Dataset} & \textbf{\#meetings} & \textbf{\#sents} & \textbf{\#names} & \textbf{\#speakers} \\ \hline
Train            & 160                 & 17,440           & 5,170            & 962                 \\ \hline
Dev              & 21                  & 1,719            & 570              & 118                 \\ \hline
Test             & 19                  & 1,562            & 429              & 106                 \\ \hline
\end{tabular}
\caption{Statistics of the dataset sampled for experiments. \textbf{\#meetings, \#sents, \#names, \#speakers} respectively denote the numbers of meetings, sentences, names, and speakers in the dataset.}
\label{tab:datasets}
\end{table}

\subsection{Hyper-parameters} We tune and select our hyper-parameters on the development set of the dataset. In particular, the models use RoBERTa large version as the PLM and are trained using the Adam optimizer with a learning rate of $1e-5$ and a batch size of $16$. All feed-forward networks have hidden vector sizes of $400$, and there are $2$ layers for the GCN. To implement the models, Pytorch version 1.7.1 \cite{NEURIPS2019_9015} and Huggingface Transformers version 3.5.1 (Apache 2.0 license) \cite{wolf-etal-2020-transformers} are used. The Trankit library version 1.0 (Apache 2.0 license) \cite{nguyen-etal-2021-trankit} is used to preprocess the data and perform named entity recognition. The model's performance is evaluated over three runs with different random seeds. Experiments are conducted on a single Tesla V100-SXM2 GPU with 32GB memory operated by Ubuntu 20.04.4 LTS.
\subsection{Evaluation Metrics} We measure the performance of the models by calculating the number of speakers that the models can successfully found their names in the transcripts. We then compute the precision, recall, and F1 scores for the models accordingly.

\begin{table}[ht]
\centering
\begin{tabular}{|l|l|l|l|}
\hline
\textbf{Models}                                            & \textbf{Precision} & \textbf{Recall} & \textbf{F1} \\ \hline
Single-name                                                & 80.3               & 50.0            & 61.6        \\ \hline
Multi-name                                                 & 78.8               & 49.1            & 60.5        \\ \hline
\begin{tabular}[c]{@{}l@{}}Multi-name\\ - GCN\end{tabular} & 75.8               & 47.2            & 58.2        \\ \hline
\end{tabular}
\caption{Performance comparision of the models on the test set of our experimental dataset.}
\label{tab:results}
\end{table}

\subsection{Results} 

Table \ref{tab:results} presents the main results of our experiments. The single-name model exhibits the highest precision at 80.3\%, suggesting that when it predicts a speaker's name, it is correct 80.3\% of the time. However, its recall is notably lower at 50.0\%, indicating that it only identifies names for half of the speakers in the dataset. The multi-name model shows a marginal decrease in precision to 78.8\% and a slight dip in recall at 49.1\%. This reduction in precision and recall may suggest that introducing multiple names into the identification process complicates the model's ability to accurately predict the correct speaker names, possibly due to the increased complexity in distinguishing between multiple speakers within the same context.

The multi-name model enhanced with Graph Convolutional Networks (GCN) further decreases in performance, with a precision of 75.8\% and a recall of 47.2\%. This result suggests that the GCN component, contrary to expectations, does not necessarily impede the model's performance. Instead, it may provide a beneficial role in the context of the multi-name setting by capturing complex patterns and relationships between speaker identities which are not as effectively discerned when the GCN is removed.

Note that, the recall scores are limited due to the fact that names of speakers are not always mentioned in the transcripts. Specifically, there are 71 speakers that we can found their names in the test transcripts, leading to the upper bound of $67.0\%$ for the recall score. This means that our best model successfully found names for $71.4\%$ of the speakers in the transcripts.

\section{Related Work}
SpeakerID is an important task for automatic organization of dialogue contents such as TV programs, radio podcasts, and online meetings and has gained significant research efforts \cite{tranter2006really,poignant2014unsupervised,poignant2015multimodal,vallet-etal-2016-speech,le2017towards}. Most previous work on SpeakerID approaches the task via multi-model setting, where the input to the models involve both videos/images and transcripts of the dialogues. Other work \cite{tranter2006really,esteve2007extracting,jousse2009automatic,kuchavrova2014study} focuses on the text-based setting, where the input to the model involves only the transcripts and text-based features of the dialogues. However, none of the previous work employs deep learning methods. Our work is the first work that employ pretrained language models \cite{devlin-etal-2019-bert,liu2019roberta} for text-based SpeakerID.

\section{Conclusions}
We proposed a novel method to automatically obtain a large-scale dataset for SpeakerID and presented novel models using pretrained language models for the task. Experimental results show that our proposed models achieve great performance, demonstrating the effectiveness and quality of our proposed dataset and model for SpeakerID. This study not only proves the practicality and effectiveness of utilizing deep learning for text-based Speaker Identification but also paves the way for further exploration in the management and retrieval of dialogue content.

\newpage

\bibliographystyle{IEEEtran}
\bibliography{anthology,custom}

\end{document}